\documentclass[letterpaper, 10 pt, conference]{ieeeconf}  

\IEEEoverridecommandlockouts                              

\usepackage{amsmath} 
\usepackage{url}
\usepackage{graphicx}
\usepackage{hyperref}

\title{\LARGE \bf
A Vision-Based Shared-Control Teleoperation Scheme for Controlling the Robotic Arm of a Four-Legged Robot
}

\author{Murilo Vinicius da Silva$^{1}$, Matheus Hipolito Carvalho$^{1}$, Juliano Negri$^{1}$, Thiago Segreto$^{1}$,\\Gustavo J. G. Lahr$^{2}$, Ricardo V. Godoy$^{1,*}$, and Marcelo Becker$^{1}$ 
\thanks{This work was carried out with the support of Petrobras, using resources from the R\&D clause of the ANP, in partnership with the University of São Paulo and the intervening foundation FAFQ, under Cooperation Agreement No. 2023/00016-6 and 2023/00013-7.}
\thanks{$^{1}$Murilo Vinicius da Silva, Matheus Hipolito Carvalho, Juliano Negri, Thiago Segreto, Ricardo V. Godoy, and Marcelo Becker are with the Department of Mechanical Engineering,
        University of São Paulo, São Carlos, Brazil
        {\tt\small becker@sc.usp.br}}%
\thanks{$^{2}$Gustavo J. G. Lahr is with the Instituto Israelita de Ensino e Pesquisa Albert Einstein, Hospital Israelita Albert Einstein, São Paulo, Brazil}
\thanks{$^{*}$Corresponding author: {\tt\small ricardo.godoy@alumni.usp.br}}
}

\begin{document}

\maketitle
\thispagestyle{empty}
\pagestyle{empty}

\begin{abstract}

In hazardous and remote environments, robotic systems perform critical tasks demanding improved safety and efficiency. Among these, quadruped robots with manipulator arms offer mobility and versatility for complex operations. However, teleoperating quadruped robots is challenging due to the lack of integrated obstacle detection and intuitive control methods for the robotic arm, increasing collision risks in confined or dynamically changing workspaces. Teleoperation via joysticks or pads can be non-intuitive and demands a high level of expertise due to its complexity, culminating in a high cognitive load on the operator. To address this challenge, a teleoperation approach that directly maps human arm movements to the robotic manipulator offers a simpler and more accessible solution. This work proposes an intuitive remote control by leveraging a vision-based pose estimation pipeline that utilizes an external camera with a machine learning-based model to detect the operator's wrist position. The system maps these wrist movements into robotic arm commands to control the robot's arm in real-time. A trajectory planner ensures safe teleoperation by detecting and preventing collisions with both obstacles and the robotic arm itself. The system was validated on the real robot, demonstrating robust performance in real-time control. This teleoperation approach provides a cost-effective solution for industrial applications where safety, precision, and ease of use are paramount, ensuring reliable and intuitive robotic control in high-risk environments.

\end{abstract}

\section{Introduction} \label{sec:introduction}
Recent advances in legged locomotion and mobile manipulation have significantly expanded the capabilities of robots in various applications, including those in challenging environments. In such scenarios, robotic systems can be employed to perform telemanipulation where direct human presence is not feasible, being employed in tasks ranging from maintenance of industrial systems~\cite{galin2019automation, Hutter2017Anymal} to search and rescue response in remote, hazardous environments~\cite{kawatsuma2012emergency, Li_Hou_Bu_Qu_2023}. The integration of robotic arms has expanded the applicability of quadruped robots in areas such as disaster response~\cite{Krotkov2017DRC}. This progress enables robots to move around and interact with the world in useful and practical ways, both navigating and physically interacting within human-centric, high-risk environments~\cite{advancing_teleop_loco}.



\begin{figure}[t!]
\centering
\includegraphics[width=\columnwidth]{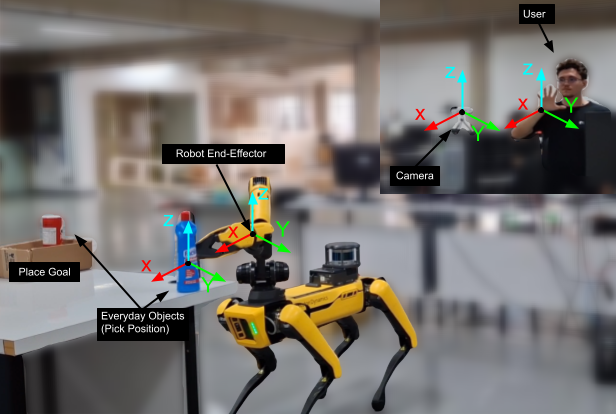} 
\caption{Experimental setup employed for testing and validating the proposed framework. The position $P_\text{wrist}$ (shown in the figure over the user's wrist) is captured by the camera, located at $P_\text{camera}$ (shown in the figure over the camera), which is employed to control the robot's end-effector position $P_\text{robot}$(shown in the figure over the robot's end-effector). The goal of the experiment is to grasp the object located at $P_\text{object}$ (shown in the figure over the everyday object) and store it at the place goal.}
\label{fig:setup}
\end{figure}


Human operators frequently manage the control of such robotic systems. Due to the complexity of these remote interactions and frameworks, a high level of training and expertise is required. Despite their sophistication, these robots often rely on classic input devices, such as joysticks or gamepads, for teleoperation. While familiar, these interfaces present steep learning curves and high cognitive loads, especially for non-experts controlling high-DoF manipulators on mobile platforms~\cite{towards_gamepad}. Managing a manipulator’s DoF motion is hard to use with joysticks~\cite{7281253}, often worsened by poor situational awareness from inadequate environmental perception~\cite{advancing_teleop_loco}. 
Operators often struggle to perceive 3D environments, plan collision-free trajectories, and execute fine motor commands under limited feedback, leading to reduced precision and fatigue~\cite{saponas2009enabling}.

To enhance intuitiveness, human-in-the-loop interfaces integrate human cognitive capabilities with robotic execution. Wearable sensors, like IMU-based motion capture systems or Electromyography (EMG) armbands, can map human movements to the robot~\cite{Mohammadi2025DisplacementMyography, Li2024EMGReview}. However, wearable devices face deployment challenges, such as time-consuming calibration and the need for intrusive physical contact. EMG systems are particularly prone to issues like motion artifacts, fatigue, electrode shifting, impedance variations, and cross-talk between adjacent muscles~\cite{multimodal_emg, affordances_emg}. 

Camera-based interfaces provide a non-invasive and accessible alternative. Such systems eliminate the need for physical instrumentation, offering a cost-effective and hands-free setup. However, such interfaces also present implementation challenges. Monocular RGB cameras lack inherent depth information, complicating 3D pose estimation. The absence of direct depth information in monocular systems results in significantly reduced accuracy compared to RGB-D approaches, with studies showing that monocular depth estimation suffers from scale ambiguity~\cite{meuleman2020single, kwon2023cameras, zhou2023disambiguating} and drift over time, particularly affecting precision in dynamic manipulation tasks~\cite{farooq2022review, chen2024d4d}. Moreover, many vision solutions are limited to specific gestures for discrete actions or use complex pipelines that hinder real-time responsiveness and generalization~\cite{rubagotti2019semi, min2025semi}. RGB-D cameras, providing direct depth measurement, simplify 3D pose estimation for manipulator control and demonstrate superior accuracy compared to monocular approaches~\cite{cao2024aberration, kwon2023cameras}.


In order to improve the robustness and applicability of human-machine interfaces, intelligent control systems can assist by partially automating task execution, reducing the overall complexity of the task, and lowering the user’s effort. Semi-autonomous systems are presented as an alternative to complex and counterintuitive manual systems, combining decoded user intentions with autonomous control modules. This work introduces a vision-based teleoperation shared-control framework designed to overcome the limitations outlined above, providing intuitive, real-time control of a quadruped's manipulator. Our system uses a depth camera for robust 3D perception to perform hand pose estimation. The proposed framework tracks the operator's wrist position for continuous end-effector control and recognizes distinct hand gestures for discrete commands, enabling natural, robust, and responsive arm teleoperation. The robot's arm follows the operator's wrist pose while continuously checking for collisions in order to ensure safe operation.

The interface enables easy switching between manual and autonomous modes via finger counting, providing a simple and low-effort method for sharing control. In autonomous mode, the robot initiates a grasp when a closed-hand gesture is detected over a detected object. The system identifies graspable items from a predefined list, selects the one with the highest confidence and closest proximity, and provides its location to help the robot locate and pick it up. This combination of depth vision, simplified hand tracking, object recognition, and intuitive mode switching aims to create a holistic, practical, and shared-control system. Importantly, it requires neither wearable devices nor traditional interfaces, addressing cost and setup barriers and offering a low-cost, accessible alternative for complex telemanipulation, especially where other systems are impractical. Experiments demonstrated a favorable balance of intuitiveness, responsiveness, and practicality for a series of pick-and-place tasks, as shown in Fig.~\ref{fig:setup}.

The rest of the paper is organized as follows: Section~\ref{sec:methods} explains how the framework was implemented and deployed. Section~\ref{sec:experiments} presents the apparatus used and the experiments performed to evaluate the framework's efficiency, along with the experimental results of this study. Section~\ref{sec:conclusion}
concludes the paper.

\section{Methods} \label{sec:methods}


\subsection{Camera-Based Motion Capture System}
\label{sec:camera-motion-capture}

\begin{figure*}[t!]
\centering
\includegraphics[width=\textwidth]{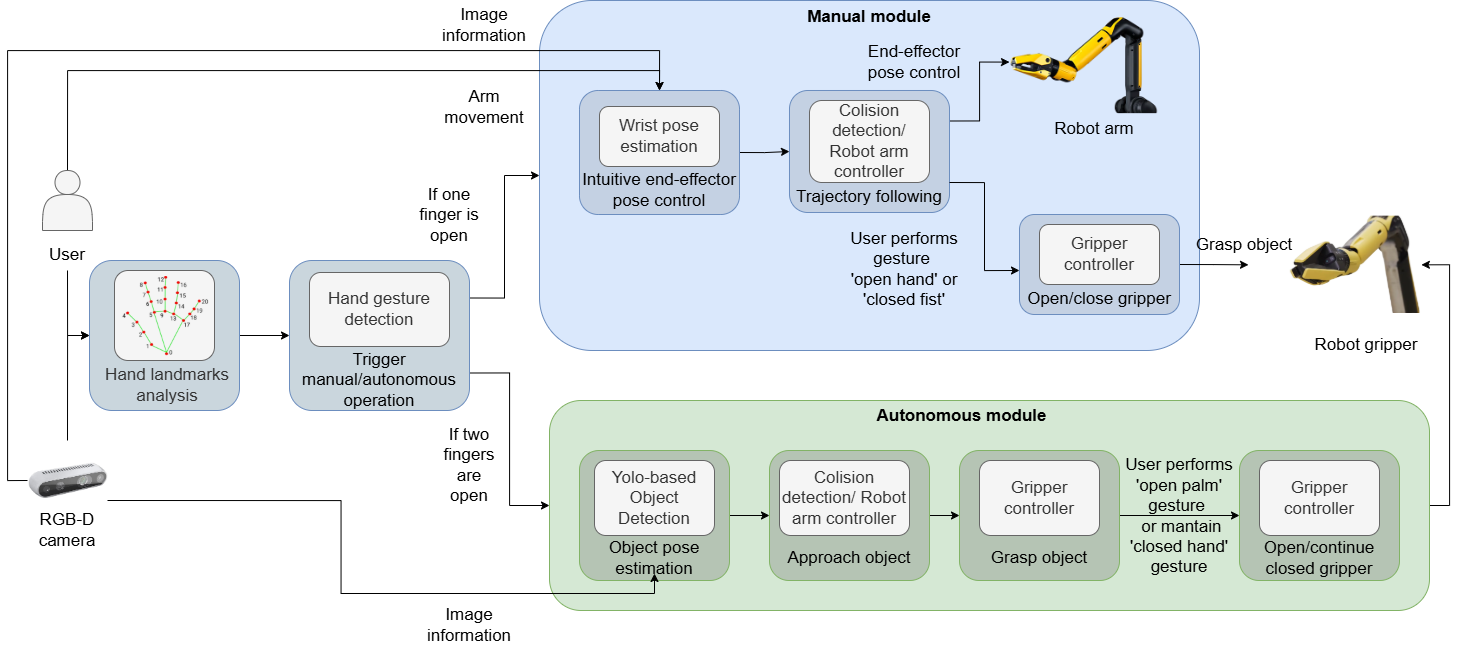} 
\caption{Overview of the proposed control pipeline, illustrating the flow from motion capture to robot actuation. A manual teleoperation module allows the user to control the robot in an intuitive manner using an external camera. A semi-autonomous module, capable of autonomously grasping an object, can be triggered by the user by closing their hands. This framework was employed to perform a pick-and-place task using different objects.}
\label{fig:control_pipeline_overview}
\end{figure*}

In order to accurately perform 3D tracking of the operator's wrist for robotic teleoperation, our system employs an Intel RealSense D435i camera to acquire synchronized RGB and depth images. The RealSense camera was chosen for its active infrared stereo technology, which provides robust depth data under varying lighting conditions. Moreover, compared to other RBG-D cameras, this may be a considerably affordable option. The camera operates at 30~Hz with a $640 \times 480$ resolution, a configuration that balances real-time responsiveness with computational efficiency for the teleoperation pipeline. Data acquisition and image alignment are managed using the \texttt{realsense-ros} package within the Robot Operating System (ROS) Noetic environment~\cite{ROS}.

The image acquisition pipeline consists of two main stages: (i) Reference Frame Calibration via ArUco Markers and (ii) Wrist Tracking via MediaPipe Pose~\cite{MediaPipe}. The experimental setup, alongside the relevant frames, are shown in Fig.~\ref{fig:setup}.

\subsubsection{Calibration}

An ArUco marker~\cite{Aruco1,Aruco2} is used to establish a stable reference frame. Upon its first detection, the system uses the camera's intrinsic parameters, provided by the RealSense ROS topic, to compute the homogeneous transformation matrix from the camera to the marker frame, $T_\text{marker}^\text{camera}$, using Eq.~\ref{eq:camera_marker}.
\begin{equation}\label{eq:camera_marker}
T_\text{marker}^\text{camera} =
\begin{bmatrix}
R & t \\
\mathbf{0} & 1
\end{bmatrix}^{-1}
\end{equation}
where $R$ is the $3\times3$ rotation matrix derived from the marker's orientation and $t$ is the $3\times1$ translation vector from the camera to the marker. This inverse transformation allows points in the camera frame to be mapped to the marker's coordinate system.

\subsubsection{Wrist Tracking}
After calibration, MediaPipe Pose is used to continuously extract the 2D pixel coordinates $(u, v)$ of the operator's right wrist from the RGB image stream. The corresponding depth value, $D_\text{raw}$, is sampled from the aligned depth image. To improve depth accuracy, we use the median value from a small $5 \times 5$ pixel window around the wrist coordinate. This 2D point with depth is then back-projected to obtain the 3D wrist position $p_\text{camera}^\text{wrist} = [X_c, Y_c, Z_c]^T$ in the camera's coordinate frame, following Eq.~\ref{eq:xyz}.

\begin{equation}
X_c = \frac{(u - c_x) \cdot Z_c}{f_x}, \quad
Y_c = \frac{(v - c_y) \cdot Z_c}{f_y}, \quad
Z_c = \frac{D_\text{raw}}{1000}
\label{eq:xyz}
\end{equation}
where $D_\text{raw}$ is in millimeters, yielding $Z_c$ in meters. To ensure robust and smooth tracking, we apply an exponential moving average filter (with $\alpha = 0.5$) to the wrist position and discard updates that exhibit a jump greater than 25\,cm between consecutive frames. The filtered 3D wrist position is finally transformed into the marker's frame using $p_\text{marker}^\text{wrist} = T_\text{marker}^\text{camera} \cdot p_\text{camera}^\text{wrist}$. This pose serves as the primary input for the control framework. Implementation details are available in our open-source repository\footnote{\url{https://github.com/EESC-LabRoM/MoveitServoing/blob/spot_teleop/}}.

\subsubsection{Hand Orientation}
\begin{figure}
    \centering
    \includegraphics[width=0.55\linewidth]{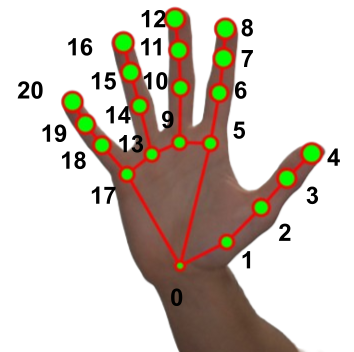}
    \caption{MediaPipe 3D landmarks numbered in order of the output. The landmarks were employed to track the user's hand in order to perform intuitive teleoperation of the robot's arm.}
    \label{fig:mediapipe-hand-landmarks}
\end{figure}

MediaPipe 3D outputs the hand landmarks, where each landmark is numbered according to Fig.~\ref{fig:mediapipe-hand-landmarks} with an index $\mathbf{p}_i$. To preprocess the data, the wrist was set as the origin, and the data was normalized by the distance between the wrist and the base of the fifth
digit, in order to make the framework more robust to different hand sizes. To do so, we use the indexes 0, 5, and 17, respectively, the wrist $\mathbf{p}_0$, the metacarpophalangeal joint of the index finger $\mathbf{p}_5$, and the metacarpophalangeal joint of the little finger $\mathbf{p}_{17}$ to compute the palm normal and the roll quaternion used by the robot using Eq.~\ref{eq:quat}, Eq.~\ref{eq:phi} and Eq.~\ref{eq:n}.

\begin{equation}
(q_x,q_y,q_z,q_w)_{\text{robot}}
=\left(-\sin\!\frac{\phi}{2},\,0,\,0,\,\cos\!\frac{\phi}{2}\right)
\label{eq:quat}
\end{equation}

\begin{equation}
\phi=\operatorname{atan2}(n_y,n_x)
\label{eq:phi}
\end{equation}

\begin{equation}
\mathbf{n}=\frac{(\mathbf{p}_{17}-\mathbf{p}_0)\times(\mathbf{p}_5-\mathbf{p}_{17})}
{\|(\mathbf{p}_{17}-\mathbf{p}_0)\times(\mathbf{p}_5-\mathbf{p}_{17})\|}
\label{eq:n}
\end{equation}

The axis remapping aligns the camera yaw to a roll about the robot $x$-axis; the quaternion is then used for end-effector orientation control and can be briefly held during grasp gestures to improve stability.

\begin{figure*}[t!]
\centering
\includegraphics[width=\textwidth]{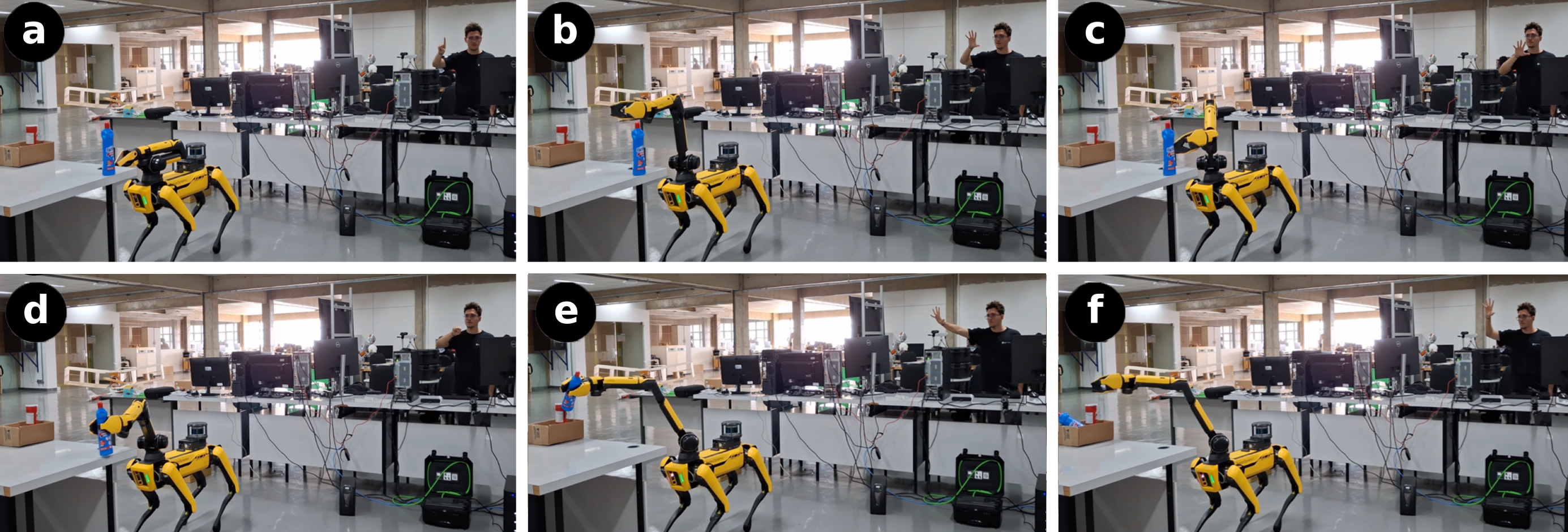}
\caption{Overview of the teleoperation framework using a robotic arm.
In a), the operator raises their index finger to the camera, activating manual control mode. b) shows the robot aligning its arm with the estimated wrist position of the user.
In c), the arm approaches the target object. The grasp is executed in d) through a closed hand gesture. In e), the operator performs an open hand gesture to initiate release, which results in the gripper opening and the object being dropped in f).}
\label{fig:teleop_sequence}
\end{figure*}

\subsection{Control Framework}\label{sec:control-framework}

The Spot robotic arm is controlled using the Boston Dynamics Software Development Kit (SDK)~\cite{BostonDynamicsSpotSDK}, ROS-based interfaces, and the MoveIt motion planning framework~\cite{MoveIt}, enabling precise end-effector pose control. Motion commands are generated based on the estimated 3D wrist pose of the user, $p_{\text{marker}}^{\text{wrist}}$, acquired through the camera-based motion capture system detailed in Section~\ref{sec:camera-motion-capture}. These commands, representing the desired end-effector pose for the robot, are transmitted to the Spot SDK via ROS, enabling real-time control of the arm. Gripper actions, such as opening and closing for object manipulation, are similarly triggered by discrete gesture-based inputs recognized by an independent gesture recognition module, which is based on hand landmarks from MediaPipe. This shared-control architecture enables intuitive teleoperation while ensuring robust and safe object manipulation. The overall control pipeline is illustrated in Fig.~\ref{fig:control_pipeline_overview}. This hybrid architecture leverages the advantages of both the proprietary Boston Dynamics SDK, which provides robust, low-level access to Spot's hardware, and the flexibility and extensibility of the open-source ROS ecosystem. ROS, together with MoveIt, manages bidirectional command and state translation, acting as the integration point for perception, planning, and actuation. 


\subsubsection{User Input Processing for Teleoperation}
The 3D wrist pose of the operator ($p_{\text{marker}}^{\text{wrist}}$), continuously provided by the motion capture system, as discussed in Section~\ref{sec:camera-motion-capture}, serves as the target end-effector pose for the Spot manipulator. This pose, originally in the ArUco marker's frame, is transformed into the robot's designated operational frame via a static calibration transformation $T_{\text{robot}}^{\text{marker}}$. The resulting target pose, $p_{\text{robot}}^{\text{target\_ee}}$, is processed by the main control node, rather than being published for consumption by other ROS modules. At system startup, a gesture recognition algorithm that counts the raised fingers based on the hand's landmarks is used to select the operation mode. When the user raises one finger, the manual teleoperation mode is triggered. When two fingers are raised and detected by the landmark-based algorithm, the semi-autonomous mode is activated. After mode selection, a separate continuous gesture classifier recognizes specific hand poses in real-time and sends action commands to the control system. Each gesture corresponds to an action depending on the current mode: i) \textit{open palm} gesture triggers the gripper to open when in manual mode or to release a grasped object when in semi-autonomous mode, while ii) \textit{closed fist} gesture closes the gripper in manual mode or initiates the grasp routine in semi-autonomous mode.

\subsubsection{Manual Teleoperation Mode}
In the primary manual teleoperation mode, the system implements a closed control loop that continuously synchronizes the Spot arm's end-effector with the operator's intended pose, as inferred from motion capture. Unlike traditional ROS-based teleoperation architectures that employ velocity streaming via MoveIt Servo, our approach leverages direct position commands using the Boston Dynamics Spot SDK. At each iteration (typically at 5\,Hz), the system queries the desired end-effector pose, as determined by the operator's wrist movement, and computes the corresponding arm command using the Spot SDK's. This pose is mapped into the robot's odometry frame, combining real-time motion capture data and ROS tf transformations. Gripper actions (open and close) are triggered in response to user gestures. The system also manages a manipulation mode for post-grasp operations. The update rate, safety interlocks, and feedback are tuned to ensure smooth and intuitive operation for real-world telemanipulation.

\subsubsection{Semi-Autonomous Grasping Mode}
The framework supports a semi-autonomous mode, triggered by operator gestures, where the robot executes grasping tasks with increased autonomy. This mode utilizes a YOLOv11-based~\cite{UltralyticsYOLO} object detection network processing images from Spot's cameras to recognize and estimate the 3D pose of manipulable objects within the scene. In this mode, the grasping routine is triggered by a specific gesture (\textit{closed fist}), and releasing a grasped object is triggered by an \textit{open palm} gesture, both recognized by the same real-time gesture classifier and mapped to their respective actions. Upon receiving the grasp-intention gesture, the system transitions to semi-autonomous control. A pre-defined or dynamically computed grasp pose for the selected object becomes the target. The robot autonomously performs the approach and gripper actuation. The system can be configured to select appropriate grasp strategies based on the object's class and geometry. The operator retains high-level control, initiating the action and having the ability to intervene. The grasping routine leverages Spot's internal API, which autonomously executes the grasp upon detection of a valid target object in the camera image. Grasp planning and execution is performed by the proprietary Spot SDK. This streamlines the pipeline but limits the ability for custom grasping strategies.

\subsubsection{Implementation Details}
The control loop, encompassing motion capture data processing, command generation, and communication with the robot, is developed for real-time performance. Perception tasks (object detection, if active) and motion planning are executed in parallel where feasible. The code for integrating user input, MoveIt functionalities, and dispatching commands to Spot, alongside Docker files, are publicly available in our repository (see footnote in Section~\ref{sec:camera-motion-capture}). Debugging overlays are provided for visualizing all perception and control stages, aiding in parameter tuning and troubleshooting.

\section{Experiments} \label{sec:experiments}

\subsection{Robotic Framework}

This work employs the Boston Dynamics Spot robot, a quadrupedal mobile platform equipped with a 6-DoF robotic arm and a jaw gripper. While the robot is capable of dynamic locomotion, this study focuses on teleoperated manipulation with the base remaining stationary. The arm offers nearly one meter of reach, enabling a wide workspace for manipulation tasks. However, as the camera does not perform pose estimation in a robust manner when the user is facing the camera backwards, the manipulation objects must be in front of the robot. Integrated sensors, including a time-of-flight (ToF) sensor and an IMU, contribute to accurate pose control. Additionally, a 4K RGB camera embedded in the gripper provides high-resolution visual feedback, supporting object localization and manipulation. The Spot platform, shown in Fig.~\ref{fig:setup}, offers a robotic platform for real-time teleoperated manipulation in real-world environments.

\subsection{Simulation Experiments}

\begin{figure}[t!]
\centering
\includegraphics[width=\columnwidth]{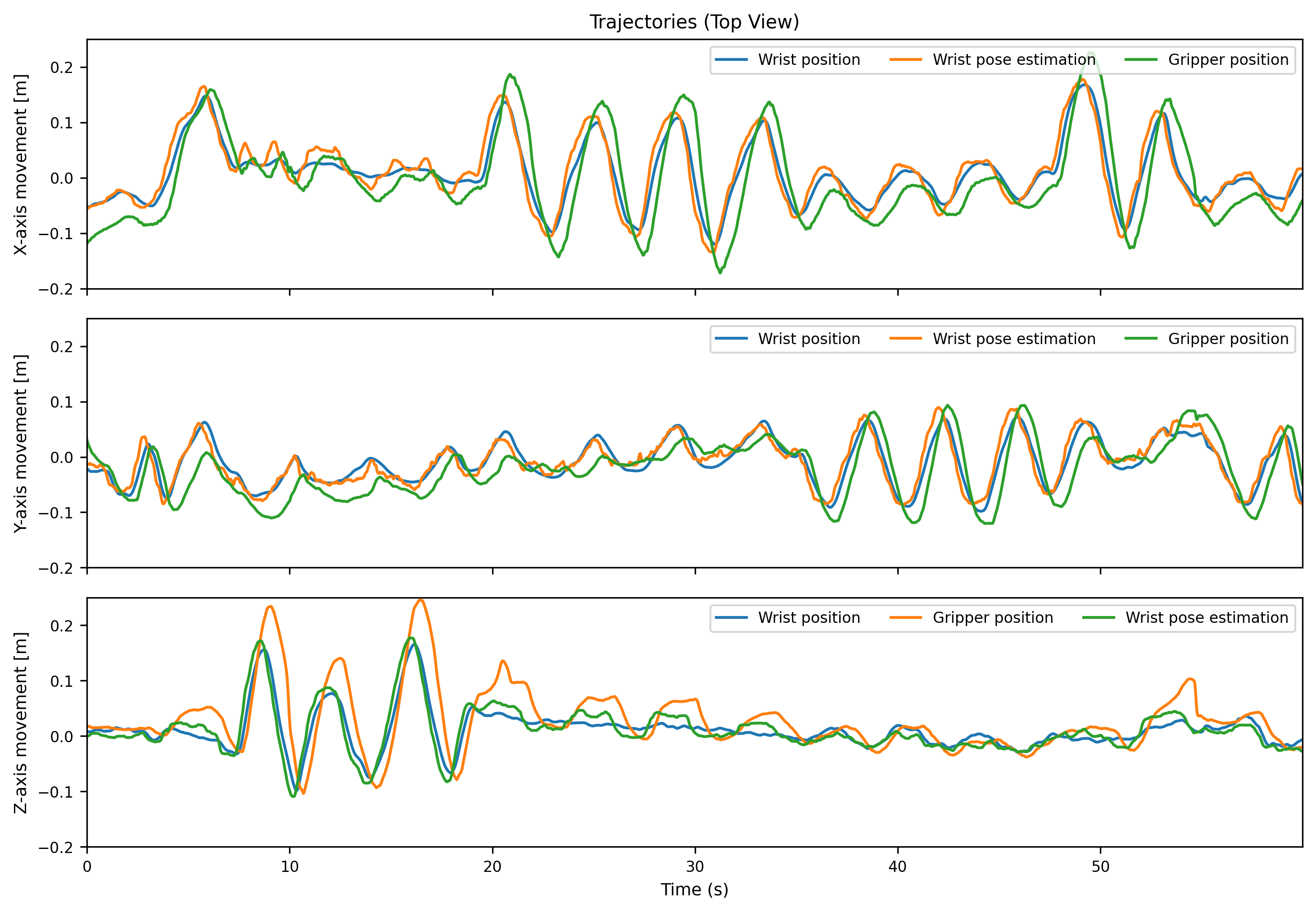} 
\caption{Decoded position of the user's wrist using the pose estimation of the wrist and using the Optitrack system, and the translated robot's end-effector position captured using the Optitrack system. These measurements were used to calculate the positional error, using the Optitrack's values as ground-truth.}
\label{fig:error}
\end{figure}

A Gazebo-based simulation environment was developed to validate the proposed teleoperation framework before real-world deployment. This simulated environment includes a virtual version of the Spot arm, integrated with MoveIt and RViz for real-time planning and collision detection. The camera-based pose estimation of the user's arm was also simulated and visualized within the environment, allowing for end-to-end testing of the control pipeline. This setup enabled assessment of the system’s precision, responsiveness, and safety mechanisms under controlled conditions, ensuring robust performance prior to real-world scenario trials.

\begin{figure}[t!]
\centering
\includegraphics[width=\columnwidth]{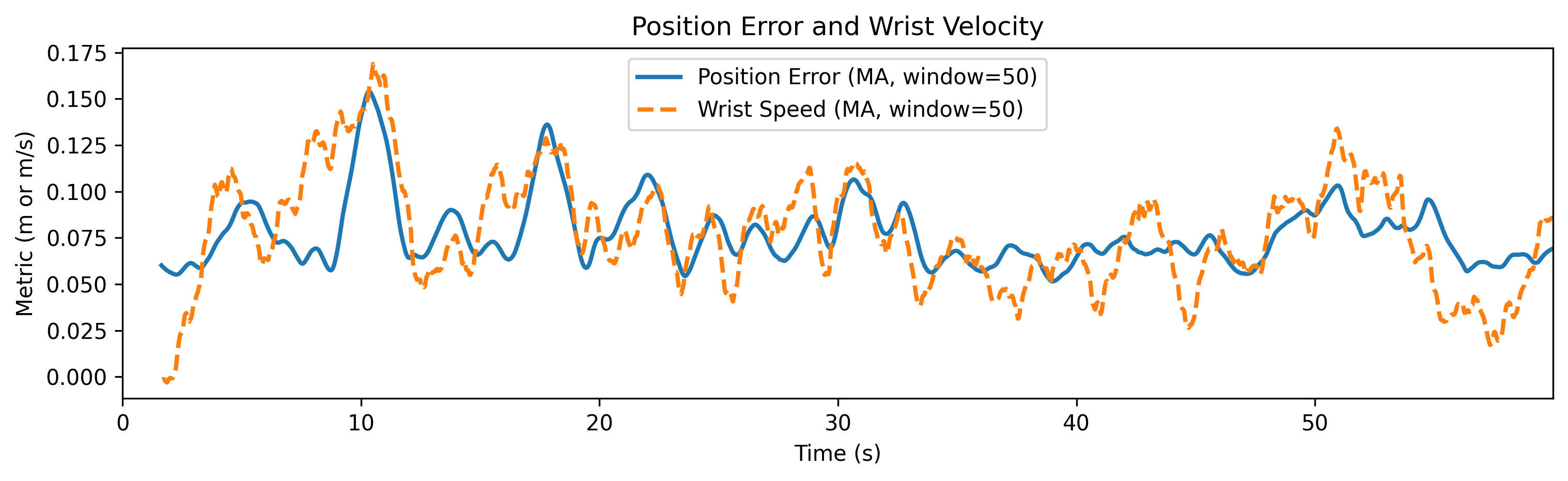} 
\caption{Positional error of the wrist's position decoded using the proposed framework and the velocity of the wrist, captured using the Optitrack system.}
\label{fig:error_vel}
\end{figure}

\subsection{Real-Time Teleoperation}

While simulation provides valuable initial insights, validating robotic frameworks in real-world scenarios is essential, particularly for teleoperation tasks where real-time responsiveness is critical to user safety and task success, preventing accidents and ensuring reliable and efficient control.

To evaluate the proposed framework, a set of real-time pick-and-place experiments was conducted by two users. In each trial, the user was asked to teleoperate the arm to pick two everyday objects and place them into a storage box, as depicted in Fig.~\ref{fig:teleop_sequence}. The experimental procedure is outlined as follows:

\begin{enumerate}
  \item \textbf{Object Approach}: The user’s wrist position and hand orientation were tracked via a vision-based system and mapped to the robot’s end-effector pose, which was sent to the SDK via ROS.
  \item \textbf{Grasp Initiation:} Upon reaching the object, the user performed a hand-closing gesture, detected by a gesture classification algorithm, to command the gripper to close.
  \item \textbf{Object Manipulation:} The user moved their arm to guide the object toward the target location, which consisted of a storage box. 
  \item \textbf{Object Release:} A hand-opening gesture triggered the gripper to open and release the object into the box, consisting of the final step of the pick-and-place task
\end{enumerate}

In the first trial, both users successfully grasped and placed a cylindrical chips can using the framework. In the second trial, a similar task was performed with a cleaning product. Successful task completion required maintaining a stable end-effector pose during grasping and adjusting gripper orientation during object placement, both achieved reliably through the proposed system. A demonstration video is available at the following URL:
\begin{center}
\href{https://ricardovgodoy.github.io/publication/2025-lars-teleop}{https://ricardovgodoy.github.io/publication/2025-lars-teleop}
\end{center}

\subsection{Precision analysis}

In order to perform a more comprehensive analysis of the proposed framework, we attached reflexive markers to the user's wrist and robot's end-effector to capture its position using an Optitrack system comprised of six Prime$^{x}$, 41 cameras and used this as ground truth to perform an analysis of the pose estimation's positional error, as the Optitrack system can achieve 3D accuracy of up to +/- 0.10 mm. The results are shown in Fig.~\ref{fig:error}. The positional error between the user's wrist from the pose estimation and the actual robot's end effector is shown in Fig.~\ref{fig:error_vel}, alongside the user's wrist velocity. A mean positional error of 0.07 m was achieved, and it can be observed that the error increases as the wrist velocity increases. This is expected, as the robot's end effector takes some time to reach the goal position, leading to this increased error.



\section{Conclusion} \label{sec:conclusion}

This paper presents an intuitive and assistive vision-based telemanipulation framework for controlling a robotic arm system of a four-legged robot. An external camera is used to track the operator's arm movement, offering intuitive control of the robotic system. The framework enables the user to control the end-effector position and orientation, where the position is obtained by tracking the user's wrist position, and orientation tracking is achieved by estimating the pose of the operator's hand. The proposed system also enables collision detection, ensuring safe operation. The framework was experimentally validated in a telemanipulation task, where two users successfully performed pick-and-place tasks in real-time. Future work will focus on implementing a path planner within the framework to achieve obstacle avoidance. Moreover, we also plan to implement a potential fields scheme~\cite{10342155} in order to improve the robustness of our proposed method through a shared-control approach, incorporating more participants, object diversity, and longer trials into the experiments. The proposed framework will be further validated by performing user and comparative studies with traditional interfaces.

\bibliographystyle{IEEEtran}
\bibliography{ref}





\end{document}